# GRAPH PARTITIONING ADVANCE CLUSTERING TECHNIQUE.

T. Soni Madhulatha

Department of Informatics, Alluri Institute of Management Sciences, Warangal, A.P.
latha.gannarapu@gmail.com

## ABSTRACT

*Clustering is a common technique for statistical data analysis, Clustering is the process of grouping the data into classes or clusters so that objects within a cluster have high similarity in comparison to one another, but are very dissimilar to objects in other clusters. Dissimilarities are assessed based on the attribute values describing the objects. Often, distance measures are used. Clustering is an unsupervised learning technique, where interesting patterns and structures can be found directly from very large data sets with little or none of the background knowledge. This paper also considers the partitioning of m-dimensional lattice graphs using Fiedler's approach, which requires the determination of the eigenvector belonging to the second smallest Eigen value of the Laplacian with K-means partitioning algorithm.*

## 1. INTRODUCTION

Unlike classification and regression, which analyze class-labeled data sets, clustering analyzes data objects without consulting class labels. In many cases, class labeled data may simply not exist at the beginning[1]. The objects are clustered or grouped based on the principle of maximizing the intraclass similarity and minimizing the interclass similarity. That is a cluster is a collection of data objects that are similar to one another within the same cluster and are dissimilar to the objects in other cluster. Each cluster so formed cab be viewed as a class of object from which rules can be derived. Besides the term data clustering as synonyms like cluster analysis, automatic classification, numerical taxonomy, botrology and typological analysis.

### 1.1 Types of Data in cluster Analysis and variable types

Before looking into the clustering algorithms first of all we need to study about the type of data that often occur in the cluster analysis. Main memory based clustering algorithms operates on either on object-by-variable structure or object-by-object structure. Object-by-variable is represented as Data matrix where as the object-by-object structure is represented as Dissimilarity matrix. As per the clustering principle we need to calculate the dissimilarity between the objects. The objects cited in data mining text book by Han and Kamber are described as:

**Interval-scaled variables**: The variables which are continuous measurements of a roughly linear scale. Example: Marks, Age, Height etc.

**Binary variables**: This variable has only two states either 0 or 1.

**Nominal variables**: Nominal is the generalization of binary variable which can take more than two states. Example rainbow colors have VIBGRO colors so six states are considered.

**Ordinal variables**: These variables are very useful for registering subjective assessment qualities that cannot be measured objectively. It is a set of continuous data of an unknown scale

**Ratio-scaled variables**: These variables make a positive measurement on a non-linear scale, such as an exponential scale.

### 1.2 Categorization of clustering methods

There exist a large number of clustering algorithms in the literature. The choice of clustering algorithm depends both on the type of data available and on the particular purpose and application. If cluster analysis is used as a descriptive or exploratory tool, it is possible to try several algorithms on the same data to see what the data may disclose. In general, major clustering methods can be classified into the following categories.

1. Hierarchical
2. Density

The above methods are not contemporary methods

Partitioning methods

3. K-means
4. K-Medoids
5. Markov **Clustering** Algorithm(MCL)
6. Non-negative matrix factorization (**NMF**)
7. Singular Value Decomposition (SVD)

The above require preliminary knowledge of data in order to choose k
Some clustering algorithms integrate the ideas of several clustering methods, so that it is sometimes difficult to classify a given algorithm as uniquely belonging to only one clustering method category.

## 2. CLASSICAL PARTITIONING METHODS

The simplest and the most fundamental version of cluster analysis is partitioning, which organizes the object of a set into several exclusive groups or clusters. The most commonly used partitioning methods are:

k-mean algorithm

k-medoids algorithm and their variations

### 2.1 K-MEANS ALGORITHM

K means clustering algorithm was developed by J. McQueen and then by J. A. Hartigan and M. A. Wong around 1975. The k-means algorithm takes the input parameter, k, and partitions a set of n objects into k clusters so that the resulting intra-cluster similarity is high whereas the inter-cluster similarity is low. Cluster similarity is measured in regard to the mean value of the objects in a cluster, which can be viewed as the cluster's center of gravity.

**Algorithm[1]:** The k-means algorithm for partitioning based on the mean value of the objects in the cluster.

Input: The number of clusters k, and a database containing n objects.

Output: A set of k clusters which minimizes the squared-error criterion.

Method:

1) arbitrarily choose k objects as the initial cluster centers;

2) repeat

3) (re)assign each object to the cluster to which the object is the most similar, based on the mean value of the objects in the cluster;

4) update the cluster means, i.e., calculate the mean value of the objects for each cluster;

5) until no change;

**Procedure**

Consider a set of objects with 2-Dimensions (PSCP and CO), let k=4 where k is number of clusters which a user would like the objects to be partitioned.

According to the algorithm we arbitrarily choose four objects as four initial cluster centers. Each object is assigned to the cluster based on the cluster center to which it is nearest. The distance between the object and cluster center is measured by Euclidean distance measure because the variables which we are using are of type of interval-based.

| Iteration0 | | |
|---|---|---|
| HALL TICKET NO. | PSCP | CO |
| 11087-i-0001 | 72 | 55 |
| 11087-i-0002 | 47 | 42 |
| 11087-i-0003 | 74 | 50 |
| 11087-i-0004 | 60 | 59 |
| 11087-i-0005 | 47 | 42 |
| 11087-i-0006 | 46 | 42 |
| 11087-i-0007 | 83 | 65 |
| 11087-i-0008 | 83 | 71 |
| 11087-i-0009 | 59 | 61 |
| **11087-i-0010** | **0** | **0** |
| 11087-i-0011 | 64 | 47 |
| 11087-i-0012 | 56 | 66 |
| 11087-i-0013 | 67 | 49 |
| 11087-i-0014 | 57 | 52 |
| 11087-i-0015 | 54 | 54 |
| 11087-i-0016 | 42 | 48 |
| 11087-i-0017 | 74 | 76 |
| 11087-i-0018 | 75 | 54 |
| 11087-i-0019 | 84 | 60 |
| 11087-i-0020 | 42 | 44 |
| 11087-i-0021 | 56 | 58 |
| 11087-i-0022 | 59 | 61 |
| 11087-i-0023 | 49 | 43 |
| **11087-i-0024** | **0** | **0** |
| 11087-i-0025 | 70 | 58 |
| 11087-i-0026 | 55 | 50 |
| 11087-i-0027 | 53 | 70 |
| 11087-i-0028 | 77 | 71 |
| 11087-i-0029 | 68 | 51 |
| 11087-i-0030 | 56 | 52 |
| 11087-i-0031 | 47 | 62 |
| 11087-i-0032 | 72 | 64 |
| 11087-i-0033 | 67 | 43 |
| **11087-i-0034** | **0** | **0** |
| 11087-i-0035 | 80 | 66 |
| **11087-i-0036** | **0** | **0** |
| 11087-i-0037 | 42 | 42 |
| 11087-i-0038 | 67 | 54 |
| 11087-i-0039 | 70 | 49 |
| 11087-i-0040 | 73 | 60 |

Table 1: Sample data points

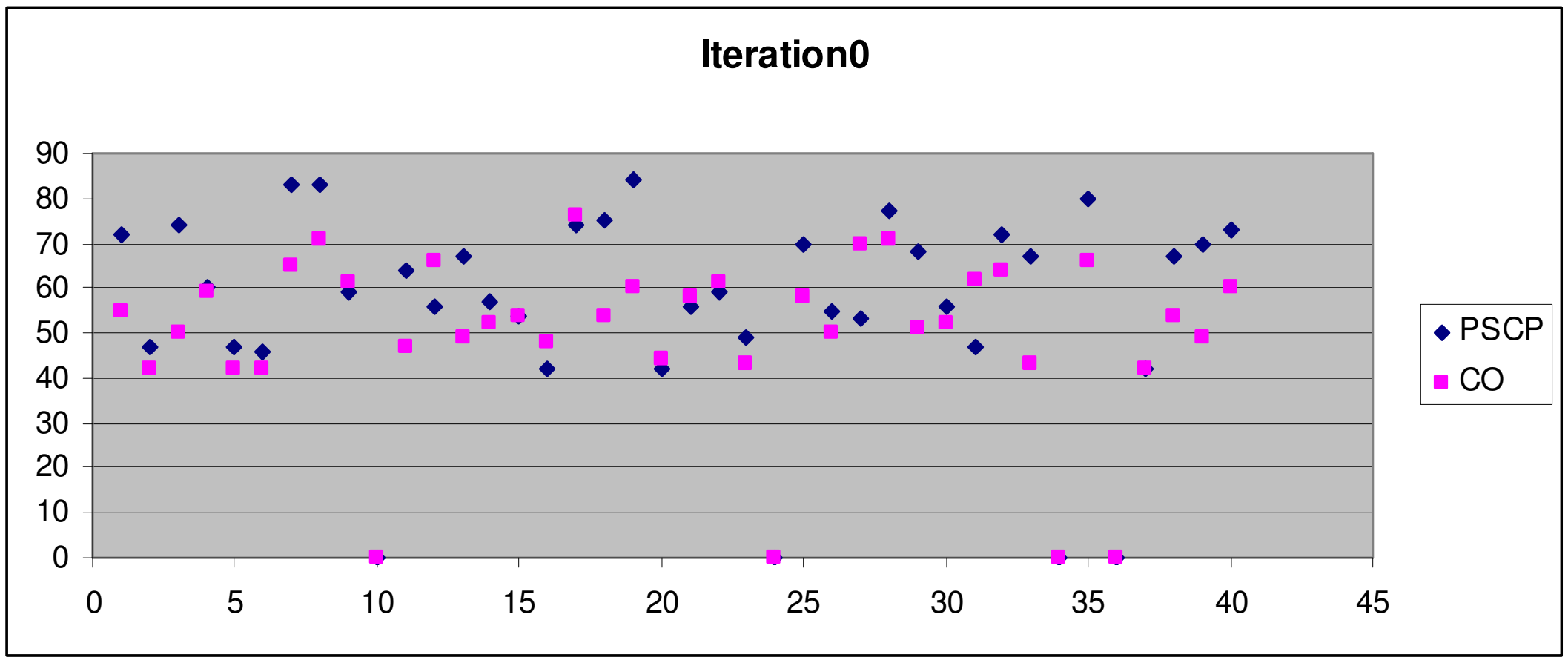

Figure:1 Initial data points distribution on XY Scatter graph

Let $c_1$, c2, c3 and c4 denote the coordinate of the cluster centers, as c1=(33,49) , c2=(68,51), c3=(75,65) and c4=(84,71). The **Euclidean distance function** measures the distance. The formula for this distance between a point ***X*** (***X1, X2,*** etc.) and a point ***Y*** (***Y1, Y2,*** etc.) is:

$$d = \sqrt{\sum_{j=1}^{n} (x_j - y_j)^2} \quad \text{------------------------------------(1)}$$

Deriving the Euclidean distance between two data points involves computing the square root of the sum of the squares of the differences between corresponding values. We can make the calculation faster by using excel function as =SQRT(SUMSQ(33-B3,49-C3)). The clusters labels used to denote the group are{1,2,3,4}. In the first Iteration we find the distances between data points and the cluster center. Now observing the column values which ever has minimum distance then under cluster group assign its label as:

| | | | Iteration1 | | | | |
|---|---|---|---|---|---|---|---|
| HALL TICKET NO. | PSCP | CO | cluster center-1 (33,49) | cluster center-2 (68,51) | cluster center-3 (75,65) | cluster center-4 (84,71) | cluster labels (1,2,3,4) |
| 11087-i-0001 | 72 | 55 | 39 | 6 | 10 | 20 | 2 |
| 11087-i-0002 | 47 | 42 | 16 | 23 | 36 | 47 | 1 |
| 11087-i-0003 | 74 | 50 | 41 | 6 | 15 | 23 | 2 |
| 11087-i-0004 | 60 | 59 | 29 | 11 | 16 | 27 | 2 |
| 11087-i-0005 | 47 | 42 | 16 | 23 | 36 | 47 | 1 |
| 11087-i-0006 | 46 | 42 | 15 | 24 | 37 | 48 | 1 |
| 11087-i-0007 | 83 | 65 | 52 | 21 | 8 | 6 | 4 |

Table 2: Distance Matrix formed in Iteration1

Next, the cluster centers are updated. That is, the mean value of each cluster is recalculated based on the current objects in the cluster. Using the new cluster centers, the objects are redistributed to the clusters based on which cluster center is the nearest. Hence after first iteration the new cluster centers are cluster center-1 (30,30), cluster center-2 (63,54),cluster center-3 (73,65) and cluster center-4 (80,71).

| | | | Iteration2 | | | | |
|---|---|---|---|---|---|---|---|
| HALL TICKET NO. | PSCP | CO | cluster center-1 (30,30) | cluster center-2 (63,54) | cluster center-3 (73,65) | cluster center-4 (80,71) | cluster labels (1,2,3,4) |
| 11087-i-0001 | 72 | 55 | 49 | 9 | 10 | 18 | 2 |
| 11087-i-0002 | 47 | 42 | 21 | 20 | 35 | 44 | 1 |
| 11087-i-0003 | 74 | 50 | 48 | 12 | 15 | 22 | 2 |
| 11087-i-0004 | 60 | 59 | 42 | 6 | 14 | 23 | 2 |
| 11087-i-0005 | 47 | 42 | 21 | 20 | 35 | 44 | 2 |
| 11087-i-0006 | 46 | 42 | 20 | 21 | 35 | 45 | 1 |
| 11087-i-0007 | 83 | 65 | 64 | 23 | 10 | 7 | 4 |
| 11087-i-0008 | 83 | 71 | 67 | 26 | 12 | 3 | 4 |
| 11087-i-0009 | 59 | 61 | 42 | 8 | 15 | 23 | 2 |
| **11087-i-0010** | **0** | **0** | 42 | 83 | 98 | 107 | 1 |
| 11087-i-0011 | 64 | 47 | 38 | 7 | 20 | 29 | 2 |
| 11087-i-0012 | 56 | 66 | 44 | 14 | 17 | 25 | 2 |
| 11087-i-0013 | 67 | 49 | 42 | 6 | 17 | 26 | 2 |
| 11087-i-0014 | 57 | 52 | 35 | 6 | 21 | 30 | 2 |

Table 3: Distance Matrix formed in Iteration2

This process of iteratively reassigning objects to clusters to improve the partitioning is referred to as iterative relocation. Eventually, no reassignment of the objects in the cluster occurs and so the process terminates. The resulting clusters are returned by clustering process.

| | | | Iteration3 | | | |
|---|---|---|---|---|---|---|
| HALL TICKET NO. | PSCP | CO | cluster center-1 (24,24) | cluster center-3 (75,59) | cluster center-4 (79,70) | cluster labels (1,2,3,4) |
| 11087-i-0001 | 72 | 55 | 57 | 5 | 17 | 3 |
| 11087-i-0002 | 47 | 42 | 29 | 33 | 43 | 2 |
| 11087-i-0003 | 74 | 50 | 56 | 9 | 21 | 3 |
| 11087-i-0004 | 60 | 59 | 50 | 15 | 22 | 2 |
| 11087-i-0005 | 47 | 42 | 29 | 33 | 43 | 2 |
| 11087-i-0006 | 46 | 42 | 28 | 34 | 43 | 2 |
| 11087-i-0007 | 83 | 65 | 72 | 10 | 6 | 4 |
| 11087-i-0008 | 83 | 71 | 75 | 14 | 4 | 4 |
| 11087-i-0009 | 59 | 61 | 51 | 16 | 22 | 2 |
| **11087-i-0010** | **0** | **0** | 34 | 95 | 106 | 1 |

Table 4: Distance Matrix formed in Iteration3

| | | | Iteration4 | | | | |
|---|---|---|---|---|---|---|---|
| HALL TICKET NO. | PSCP | CO | cluster center-1 (0,0) | cluster center-2 (56,52) | cluster center-3 (75,59) | cluster center-4 (79,70) | cluster labels (1,2,3,4) |
| 11087-i-0001 | 72 | 55 | 91 | 16 | 5 | 17 | 3 |
| 11087-i-0002 | 47 | 42 | 63 | 13 | 33 | 43 | 2 |
| 11087-i-0003 | 74 | 50 | 89 | 18 | 9 | 21 | 3 |
| 11087-i-0004 | 60 | 59 | 84 | 8 | 15 | 22 | 2 |
| 11087-i-0005 | 47 | 42 | 63 | 13 | 33 | 43 | 2 |
| 11087-i-0006 | 46 | 42 | 62 | 14 | 34 | 43 | 2 |
| 11087-i-0007 | 83 | 65 | 105 | 30 | 10 | 6 | 4 |
| 11087-i-0008 | 83 | 71 | 109 | 33 | 14 | 4 | 4 |
| 11087-i-0009 | 59 | 61 | 85 | 9 | 16 | 22 | 2 |
| **11087-i-0010** | **0** | **0** | 0 | 76 | 95 | 106 | 1 |

Table 5: Distance Matrix formed in Iteration4

| | | | Iteration7 | | | | |
|---|---|---|---|---|---|---|---|
| HALL TICKET NO. | PSCP | CO | cluster center-1 (0,0) | cluster center-2 (52,53) | cluster center-3 (70,52) | cluster center-4 (79,68) | cluster labels (1,2,3,4) |
| 11087-i-0001 | 72 | 55 | 91 | 20 | 4 | 15 | 3 |
| 11087-i-0002 | 47 | 42 | 63 | 12 | 25 | 41 | 2 |
| 11087-i-0003 | 74 | 50 | 89 | 22 | 4 | 19 | 3 |
| 11087-i-0004 | 60 | 59 | 84 | 10 | 12 | 21 | 2 |
| 11087-i-0005 | 47 | 42 | 63 | 12 | 25 | 41 | 2 |
| 11087-i-0006 | 46 | 42 | 62 | 13 | 26 | 42 | 2 |
| 11087-i-0007 | 83 | 65 | 105 | 33 | 18 | 5 | 4 |
| 11087-i-0008 | 83 | 71 | 109 | 36 | 23 | 5 | 4 |
| 11087-i-0009 | 59 | 61 | 85 | 11 | 14 | 21 | 2 |
| **11087-i-0010** | **0** | **0** | 0 | 74 | 87 | 104 | 1 |
| 11087-i-0011 | 64 | 47 | 79 | 13 | 8 | 26 | 3 |
| 11087-i-0012 | 56 | 66 | 87 | 14 | 20 | 23 | 2 |
| 11087-i-0013 | 67 | 49 | 83 | 16 | 4 | 22 | 3 |
| 11087-i-0014 | 57 | 52 | 77 | 5 | 13 | 27 | 2 |
| 11087-i-0015 | 54 | 54 | 76 | 2 | 16 | 29 | 2 |

Table 6: Distance Matrix formed in Iteration7

The k-means algorithm is applied on the 40 records with two attributes and to obtain final result, the algorithm under goes seven iteration and stops. After Iteration7 we see that the new mean values obtained for four clusters are the same as that of previous step.

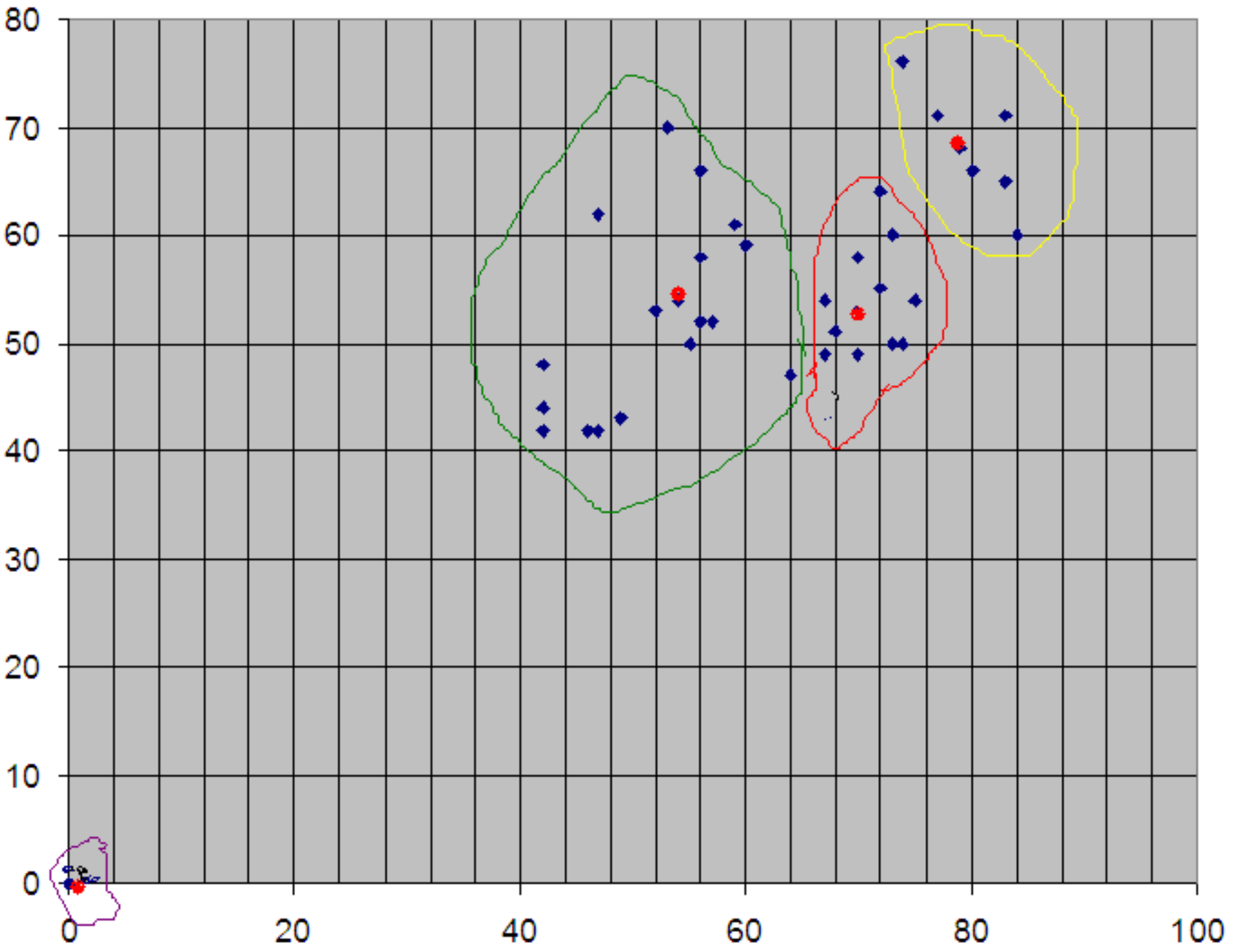

Figure 2: Final object distribution among four clusters

Instead of considering four clusters, if we consider six clusters (k=6) the algorithm iterates for two times and terminates as the mean values never change after it.

| Iteration-1 | | | | | | | | | |
|---|---|---|---|---|---|---|---|---|---|
| HALL TICKET NO. | PSCP | CO | cluster center 1(0,0) | cluster center 2(42,42) | cluste r center 3(54, 54) | cluster center 4(68,5 1) | cluster center 5(77,71 ) | cluster center 6(83,65 ) | cluster group (1,2,3, 4,5,6) |
| 11087-i-0001 | 72 | 55 | 91 | 33 | 18 | 6 | 17 | 15 | 4 |
| 11087-i-0002 | 47 | 42 | 63 | 5 | 14 | 23 | 42 | 43 | 2 |
| 11087-i-0003 | 74 | 50 | 89 | 33 | 20 | 6 | 21 | 17 | 4 |
| 11087-i-0004 | 60 | 59 | 84 | 25 | 8 | 11 | 21 | 24 | 3 |
| 11087-i-0005 | 47 | 42 | 63 | 5 | 14 | 23 | 42 | 43 | 2 |
| 11087-i-0006 | 46 | 42 | 62 | 4 | 14 | 24 | 42 | 44 | 2 |
| 11087-i-0007 | 83 | 65 | 105 | 47 | 31 | 21 | 8 | 0 | 6 |
| 11087-i-0008 | 83 | 71 | 109 | 50 | 34 | 25 | 6 | 6 | 6 |
| 11087-i-0009 | 59 | 61 | 85 | 25 | 9 | 13 | 21 | 24 | 3 |
| **11087-i-0010** | **0** | **0** | 0 | 59 | 76 | 85 | 105 | 105 | 1 |

Table 7: Distance Matrix formed in Iteration1 for six clusters

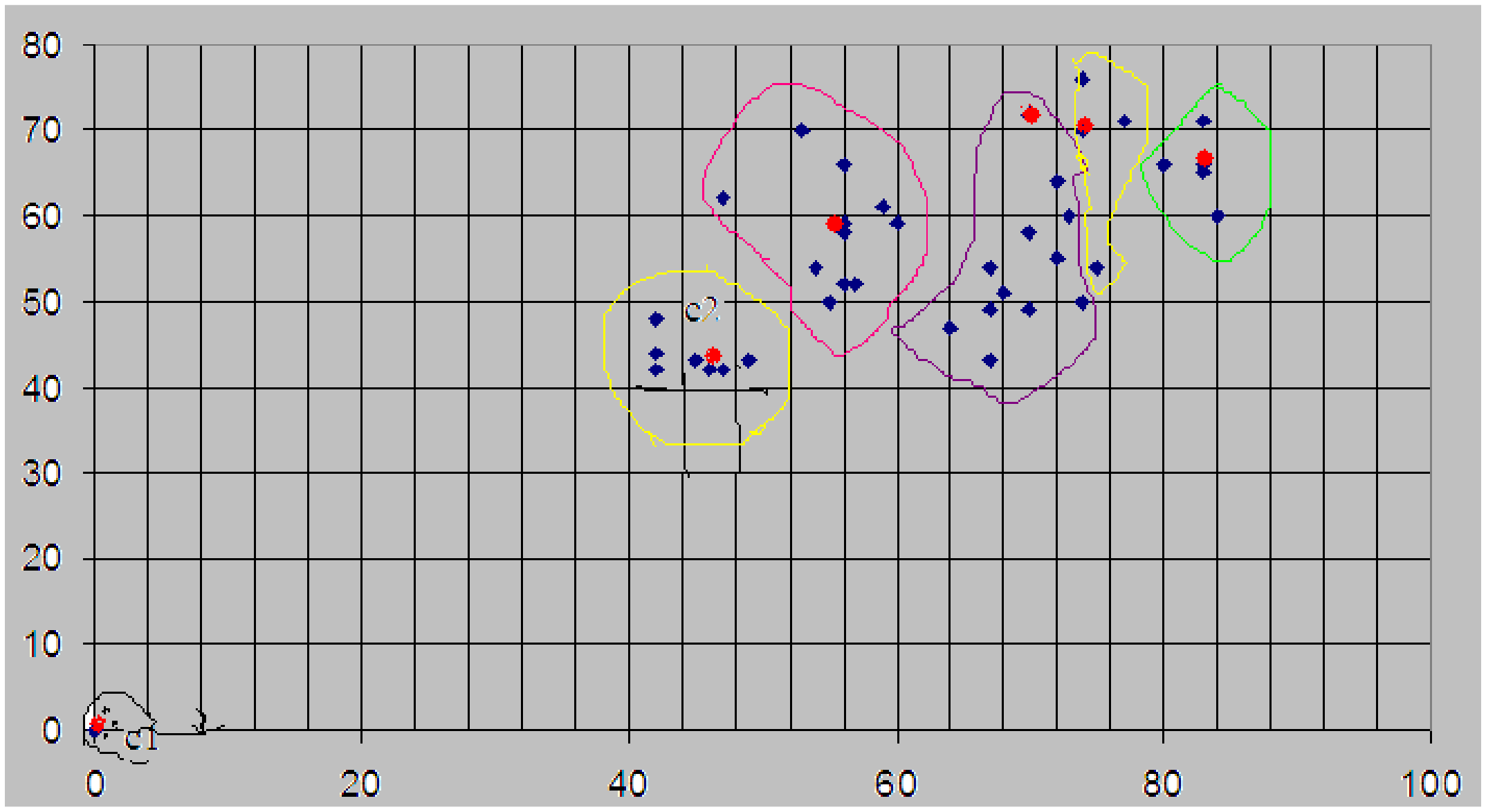

Figure 3: Object distribution among six clusters

**K**-Means method is not guaranteed to converge to global optimum and often terminates at a local optimum. To obtain good results in practice it is common to run the k-means algorithm multiple times with different initial clusters.

**Advantages and Disadvantages of K-Means**

- K-means is relatively scalable and efficient in processing large datasets.
- The method is not suitable for discovering clusters with non-convex shapes or of very different sizes.
- It is sensitive to noise and outlier data

## 3. FIEDLER'S METHOD

Fiedler's approach to clustering, which theoretically determines the relation between the size of the obtained clusters and the number of links that are cut by this partitioning as a function of a threshold $\alpha$ and of graph properties such as the number of nodes and links. When applying Fiedler's beautiful results to the Laplacian matrix Q of a graph, the eigenvector belonging to the second smallest eigenvalue, known as the algebraic connectivity, needs to be computed.
Finding the laplacian matrix requires construction of A adjacency matrix, and D degree matrix, So the Laplacian matrix L is formed as:

L = D – A --------------------------------------(2)

Given a simple graph *G* with *n* vertices, its laplacian matrix is defined as:

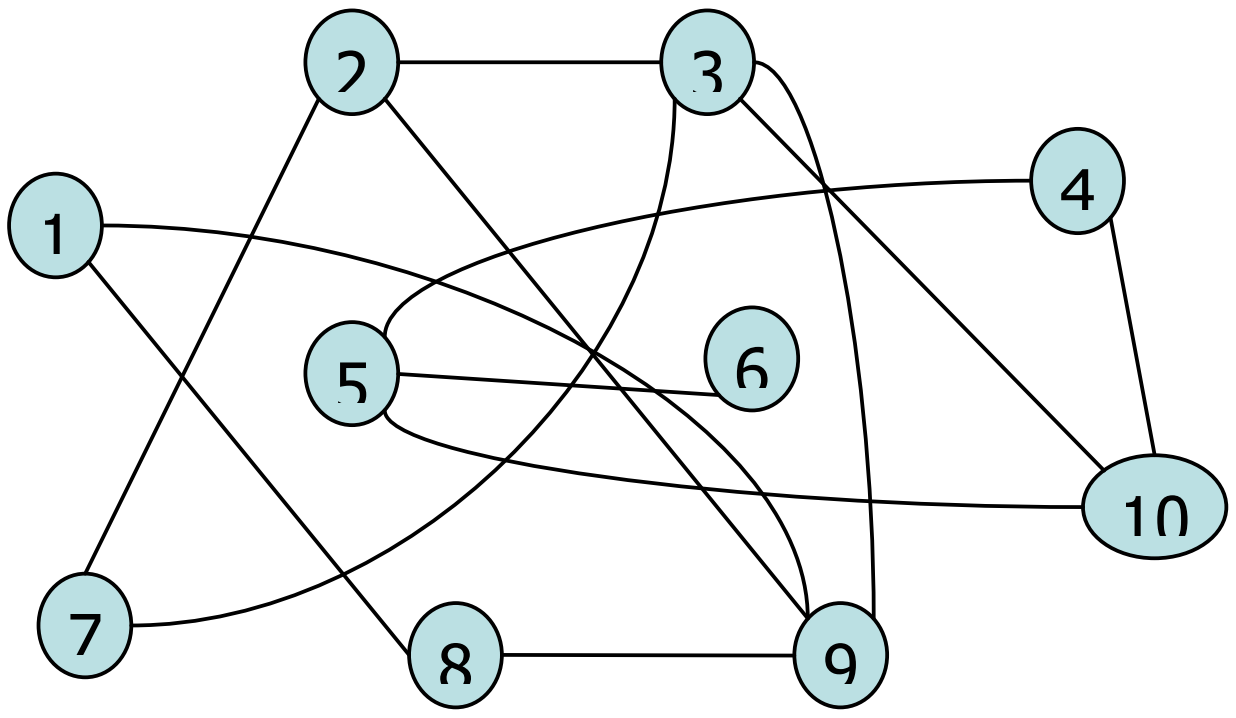

Figure 4: Example graph

### 3.1 Adjacency Matrix

The adjacency matrix of a finite graph G of n vertices is the n × n matrix where the non-diagonal entry $a_{ij}$ is the number of edges from vertex i to vertex j, and the diagonal entry $a_{ii}$, is either once or twice the number of edges from vertex i to itself.

$$A = \begin{bmatrix} 0 & 0 & 0 & 0 & 0 & 0 & 0 & 1 & 1 & 0 \\ 0 & 0 & 1 & 0 & 0 & 0 & 1 & 0 & 1 & 0 \\ 0 & 1 & 0 & 0 & 0 & 0 & 1 & 0 & 1 & 1 \\ 0 & 0 & 0 & 0 & 1 & 1 & 0 & 0 & 0 & 1 \\ 0 & 0 & 0 & 1 & 0 & 1 & 0 & 0 & 0 & 1 \\ 0 & 0 & 0 & 1 & 1 & 0 & 0 & 0 & 0 & 1 \\ 0 & 1 & 1 & 0 & 0 & 0 & 0 & 0 & 0 & 0 \\ 1 & 0 & 0 & 0 & 0 & 0 & 0 & 0 & 1 & 0 \\ 1 & 1 & 1 & 0 & 0 & 0 & 0 & 1 & 0 & 0 \\ 0 & 0 & 1 & 1 & 1 & 1 & 0 & 0 & 0 & 0 \end{bmatrix}$$

Figure 5: The adjacency matrix for the graph-1

### 3.2 Degree matrix

In the mathematical field of graph theory the degree matrix is a diagonal matrix which contains information about the degree of each vertex. That is the count of edges connecting a vertex v. If $i \neq j$ then replace the cell value with 0 other wise degree of the vertex $v_i$

$$D = \begin{bmatrix} 2 & 0 & 0 & 0 & 0 & 0 & 0 & 0 & 0 & 0 \\ 0 & 3 & 0 & 0 & 0 & 0 & 0 & 0 & 0 & 0 \\ 0 & 0 & 4 & 0 & 0 & 0 & 0 & 0 & 0 & 0 \\ 0 & 0 & 0 & 3 & 0 & 0 & 0 & 0 & 0 & 0 \\ 0 & 0 & 0 & 0 & 3 & 0 & 0 & 0 & 0 & 0 \\ 0 & 0 & 0 & 0 & 0 & 3 & 0 & 0 & 0 & 0 \\ 0 & 0 & 0 & 0 & 0 & 0 & 2 & 0 & 0 & 0 \\ 0 & 0 & 0 & 0 & 0 & 0 & 0 & 2 & 0 & 0 \\ 0 & 0 & 0 & 0 & 0 & 0 & 0 & 0 & 4 & 0 \\ 0 & 0 & 0 & 0 & 0 & 0 & 0 & 0 & 0 & 4 \end{bmatrix} -$$

Figure 6: The degree matrix of graph-1

### 3.3 Laplacian matrix

Given a simple graph G with n vertices, its Laplacian matrix is defined as:
L(i,j)=degree of vertex $v_i$ if i=j, if i≠j and $v_i$ is not adjacent to $v_j$ and in all other case fill it with 0.

$$L=\begin{bmatrix} 2 & 0 & 0 & 0 & 0 & 0 & 0 & -1 & -1 & 0 \\ 0 & 3 & -1 & 0 & 0 & 0 & -1 & 0 & -1 & 0 \\ 0 & -1 & 4 & 0 & 0 & 0 & -1 & 0 & -1 & -1 \\ 0 & 0 & 0 & 3 & -1 & -1 & 0 & 0 & 0 & -1 \\ 0 & 0 & 0 & -1 & 3 & -1 & 0 & 0 & 0 & -1 \\ 0 & 0 & 0 & -1 & -1 & 3 & 0 & 0 & 0 & -1 \\ 0 & -1 & -1 & 0 & 0 & 0 & 2 & 0 & 0 & 0 \\ -1 & 0 & 0 & 0 & 0 & 0 & 0 & 2 & -1 & 0 \\ -1 & -1 & -1 & 0 & 0 & 0 & 0 & -1 & 4 & 0 \\ 0 & 0 & -1 & -1 & -1 & -1 & 0 & 0 & 0 & 4 \end{bmatrix}$$

Figure 7: the laplacian matrix for graph -1

### 3.4 Fiedler method

This method partitions the data set S into two sets S1 and S2 based on the eigen Vector V corresponding to the 2nd smallest eigen value of laplacian matrix. Consider the Equations

$$y_i = \sum_{k=1}^{n} a_{j_k} x_k \qquad j = 1........,n \text{ --------------- } (3)$$

Represent a linear transformation from the variables $x_1, x_2, ............. x_n$ to the variables y1,...........$y_n$ ; we can write this in matrix notation as Y=AX, where Y is a column vector and A=($a_{ij}$) is matrix transformation. In many situations, we need to transform a vector into a scalar multiple of itself.

i.e. AX=λX-----------------------------------------------(4) where λ is a scalar.

Such problems are known as eign value problems. Let A be an n x n symmetric matrix and x is known as eigen vector corresponding to the eigen values. To obtaine eigen vector we need to solve (A-λI)x=0. x=0 is a trivial solution of this linear system for any λ. For the system to have a non-trivial solution, the matrix A-λI must be singular. The scalar λ and the non-zero vector x satisfying(4) exist if |A-λI|=0.

$$p_n(\lambda) = \begin{vmatrix} a_{11}-\lambda & a_{12} \ldots\ldots\ldots\ldots\ldots\ldots a_{1n} \\ a_{21} & a_{22}-\lambda \ldots\ldots\ldots\ldots a_{2n} \\ . & \\ . & \\ . & \\ a_{n1} & a_{n2} \ldots\ldots\ldots\ldots\ldots\ldots a_{nn}-\lambda \end{vmatrix} = 0.$$

By expansion of this determinant we get an nth degree polynomial in λ and $p_n(\lambda)$ is known as the characteristic polynomial of A and $p_n(\lambda)=0$ as the characteristic equation of A. $p_n(\lambda)=0$ has n roots, which may be real or complex. The roots are the eigen values of the matrix.
As per Rayleigh Quotient Theorem Solution:

$-\lambda_1$=0, the smallest right-hand eigenvalue of the *symmetric matrix*, L

$-\lambda_1$ corresponds to the trivial eigenvector

$v_1$= e = [1, 1, …, 1].

Based on a symmetric matrix, L, we search for the eigenvector, $v_2$, which is furthest away from e. Now $v_2$ gives relation information about the nodes. This relation is usually decided by separating the values across zero.
A theoretical justification is given by Miroslav Fiedler. Hence, $v_2$ is called the Fiedler vector. Hence $v_2$ is used to recursively partition the graph by separating the components into negative and positive values.
Entire Graph: sign(V)=[ +, +, +, -, -, -, +, +, +, -]
1 2 3 4 5 6 7 8 9 10

Iteration 1:
Positives = [+, +, +, +, +, +]
1 2 3 7 8 9
Negatives= [-, -, -, -]
4 5 6 10

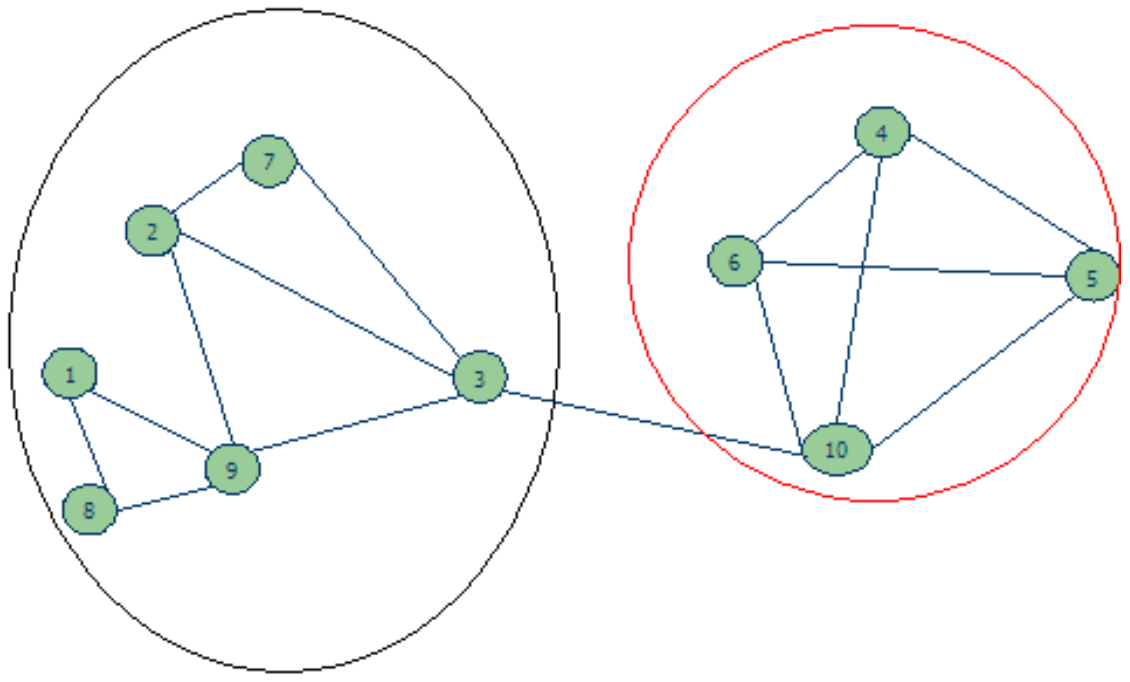

Figure 12: Graph-1 after first iteration.

Sign(v2)=[+, -, -, -, +, +, ]
1 2 3 7 8 9
Iteration 2:
Positives =[+, +, + ]
1 8 9
Negatives =[-, -, - ]
1 3 7

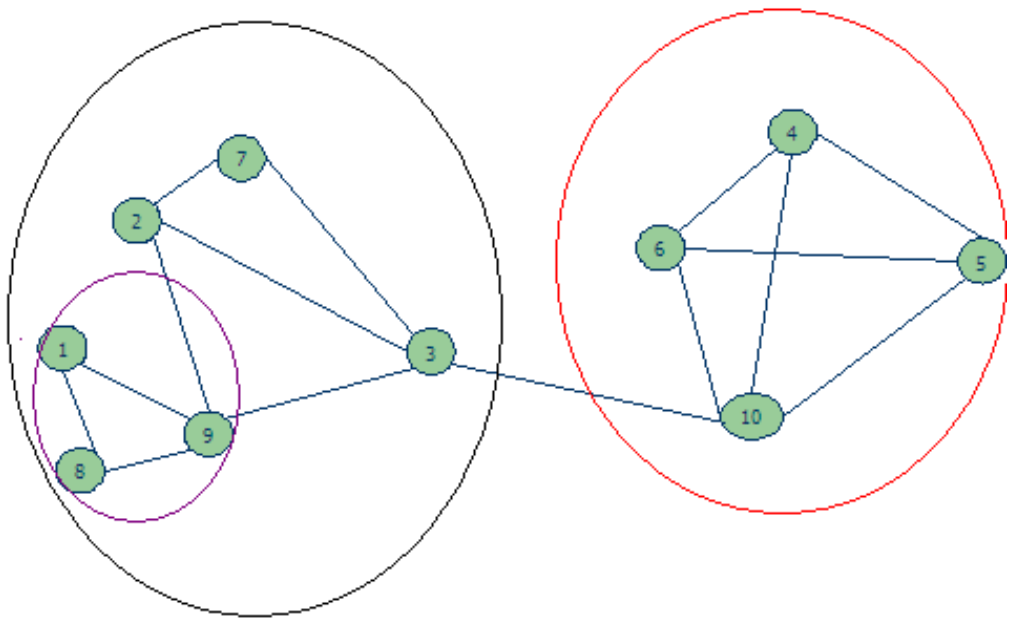

Figure 13: Graph after iteration -2

Sign(v3)= [ +,+, - ]
1 8 9

## 4. METHODOLOGY OF EXPERIMENTATION

We observed at several different eigenvectors, followed the Fiedler algorithm and then coded in Matlab using eigs(), eig() by taking small samples having known clusters.

**MATLAB CODE:**
**Steps:**

1. Enter the Laplacian matrix in matlab as:

```
a=[2 0 0 0 0 0 0 -1 -1 0;0 3 -1 0 0 0 -1 0 -1 0; 0 -1 4 0 0 0 -1 0 -1 -1;0 0 0 3 -1 -1 0 0 0 -1;0 0 0 -1 3 -1 0 0 0 -1;0 0 0 -1 -1 3 0 0 0 -1;0 -1 -1 0 0 0 2 0 0 0; -1 0 0 0 0 0 0 2 -1 0;-1 -1 -1 0 0 0 0 -1 4 0;0 0 -1 -1 -1 -1 0 0 0 4]

a =
     2     0     0     0     0     0     0    -1    -1     0
     0     3    -1     0     0     0    -1     0    -1     0
     0    -1     4     0     0     0    -1     0    -1    -1
     0     0     0     3    -1    -1     0     0     0    -1
     0     0     0    -1     3    -1     0     0     0    -1
     0     0     0    -1    -1     3     0     0     0    -1
     0    -1    -1     0     0     0     2     0     0     0
    -1     0     0     0     0     0     0     2    -1     0
    -1    -1    -1     0     0     0     0    -1     4     0
     0     0    -1    -1    -1    -1     0     0     0     4
```

2. Find the eign values from eign vector

```
>> eig(a)
[V D]=eigs(a, 2, 'SA');

ans =
    0.0000
    0.2602
    0.8638
    3.0000
    3.0607
    4.0000
    4.0000
    4.0000
    5.0000
    5.8154
```

3. Display the second smallest of Laplacian matrix

```
D(2,2)
ans =     0.2602
```

4. The sign obtained for the entire graph is

sign(V)=[ +, +, +, -, -, -, +, +, +, -]
1 2 3 4 5 6 7 8 9 10

**Iteration-2**

a=[2 0 0 0 -1 -1;0 3 -1 -1 0 -1;0 -1 4 -1 0 -1;0 -1 -1 2 0 0;-1 0 0 0 2 -1;-1 -1 -1 0 -1 4]
[V D]=eigs(a,2,'SA');
D(2,2)
ans = *0.8591*

sign(V)= = [+, +, +, +, +, +]
1 2 3 7 8 9

The final graph obtained for the graph-1

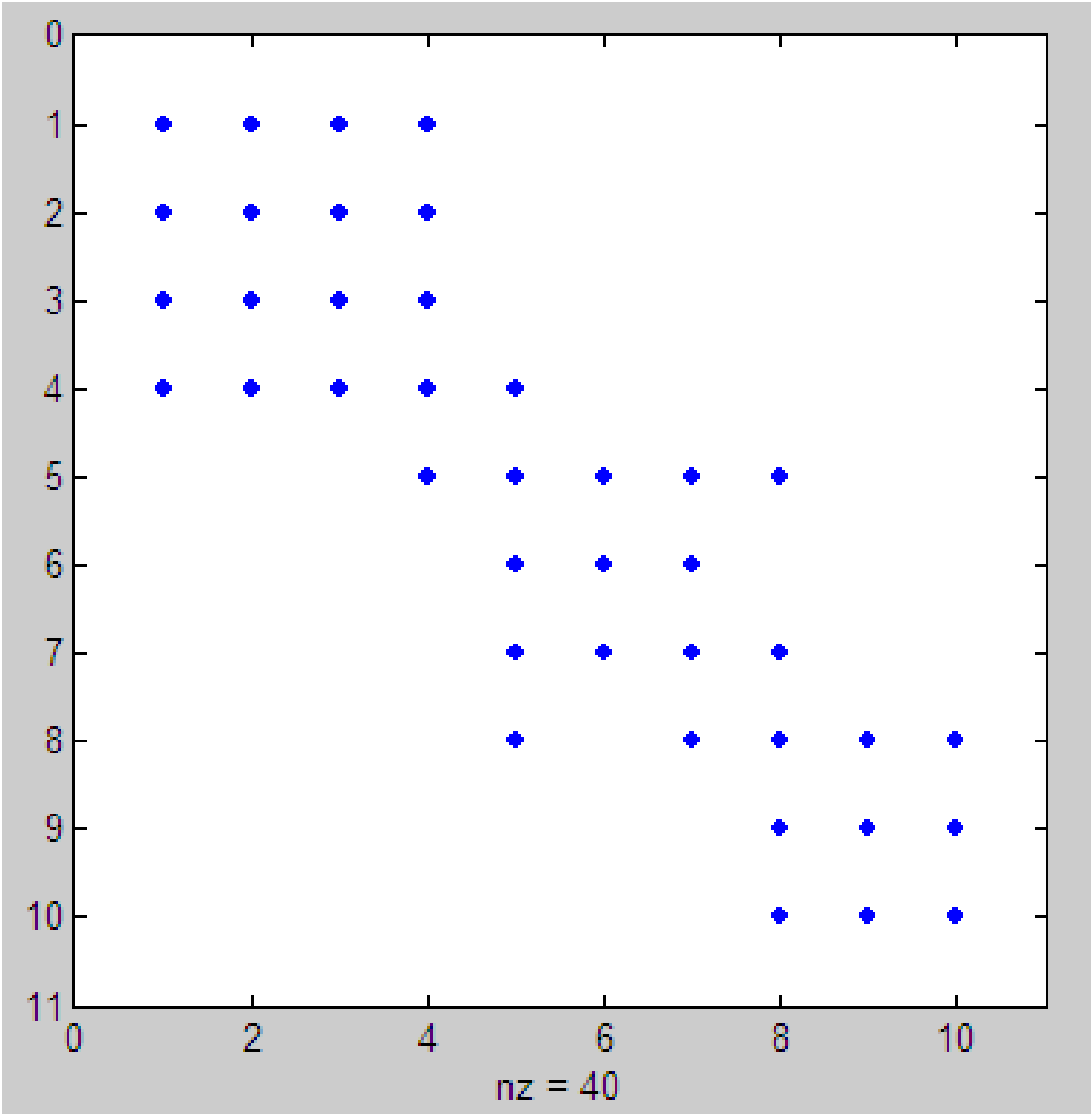

Figure 14: Plot-graph for the graph-1

## 4. CONCLUSION

The advantage of the K-Means algorithm is its favorable execution time. Its drawback is that the user has to know in advance how many clusters are searched for. It is observed that K-Means algorithm is efficient for smaller data sets only. Fielder's method doesn't require the preliminary knowledge of the number of clusters, but most clustering methods require matrices to be symmetric. Symmetrizing techniques either distort original information or greatly increase the size of the dataset moreover there are many applications where the data is not symmetric like Hyperlinks on the Web.

### Author

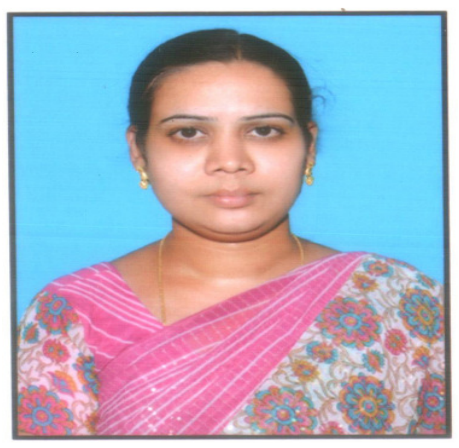

**T. Soni Madhulatha** obtained MCA from Kakatiya University in 2003 and M. Tech (CSE) from JNTUH in 2010.She has 8 years of teaching experience and she is presently working as Associate professor in Department of Informatics in Alluri Institute of Management Sciences, Hunter Road Warangal. She published papers in various National and International Journals and Conferences. She is a Life Member of ISTE , IAENG and APSMS.